\documentclass[11pt]{article}

\usepackage[preprint]{acl}

\usepackage{times}
\usepackage{latexsym}

\usepackage[T1]{fontenc}

\usepackage[utf8]{inputenc}

\usepackage{microtype}

\usepackage{inconsolata}

\usepackage{graphicx}
\usepackage{amssymb}
\usepackage{listings}
\usepackage[table,xcdraw]{xcolor}
\usepackage{booktabs}
\usepackage{adjustbox}

\usepackage{multirow}

\usepackage[most]{tcolorbox}
\usepackage{subcaption}
\usepackage{enumitem}

\usepackage{makecell}
\usepackage{float}

\lstdefinelanguage{json}{
    basicstyle=\ttfamily\small,
    breaklines=true,
    string=[s]{"}{"},
    comment=[l]{//},
    morestring=[b]',
}

\title{Modeling Student Sensemaking \\ with LLMs and Knowledge‑Graph‑Guided Inference}

\author{
  \textbf{Özge Alacam\textsuperscript{1}} 
  \textbf{Zübeyde Demet Kirbulut Güneş\textsuperscript{2}}, \\
  \textbf{Funda Ekici\textsuperscript{2}},
  \textbf{Nurcan Turan-Oluk\textsuperscript{2}},
  \textbf{Dilay Dinçdemir\textsuperscript{2}}, \\
  \textbf{Hakkı Kadayıfçı\textsuperscript{2}},
  \textbf{Sevinç Nihal Yeşiloğlu\textsuperscript{2}},
  \textbf{Burcu Işık\textsuperscript{2}},
  \textbf{Halil Tümay\textsuperscript{2}},
  \textbf{Sinem Gencer\textsuperscript{2}}
\\
  \textsuperscript{1}Center for Information and Language Processing, LMU München, Munich, Germany \\
  \textsuperscript{2}Department of Chemistry Education, Gazi Faculty of Education, Gazi University, Ankara, Türkiye
\\
  \small{
    \textbf{Correspondence:} \href{mailto:ozge.alacam@lmu.de}{oezge.alacam@lmu.de}, \href{mailto:zdgunes@gazi.edu.tr}{zdgunes@gazi.edu.tr}
  }
}

\begin{document}
\maketitle 

\begin{abstract}
Collaborative science learning requires nuanced interpretation of student dialogue to characterize how learners identify knowledge gaps, build explanations, and work toward resolution — a theory-driven analysis that is labor-intensive and difficult to scale. We investigate whether instruction-tuned large language models (LLMs) can support multidimensional analysis of collaborative sensemaking without task-specific training, and whether structured knowledge-state information improves model inference. We evaluate two mid-size LLMs on 23 richly annotated, expert-labeled episodes across prompting conditions that vary definitional scaffolding, reasoning mode, and turn structure. Without reasoning, models tend to overpredict successful sensemaking; reasoning-enabled prompting improves identification of unsuccessful cases. Knowledge-state diagnostics provide additional grounding, improving detection of unsuccessful sensemaking and increasing agreement with expert annotations. No single configuration performs best across all sensemaking dimensions, underscoring the multidimensional nature of the task.
\end{abstract}

\section{Introduction}
Collaborative sensemaking in science learning examines how learners identify and work to resolve gaps or inconsistencies in their understanding through iterative explanation building \cite{hunter2021making, odden2019defining, odden2019vexing, odden2018sensemaking}. Manually analyzing collaborative student dialogue for sensemaking is labor-intensive: it requires domain experts to identify not only episode-level outcomes, but also theoretically grounded process features - including focal knowledge gaps, epistemic emotions, and components of explanation building. These demands make sensemaking analysis difficult to automate reliably. Unlike standard text classification, sensemaking labels are not derivable from surface-level lexical cues;  they depend on the epistemic function of utterances, the structure of explanation building, and the resolution of a problem that may never be explicitly stated.

Instruction-tuned LLMs offer a potential means of supporting this analysis without task-specific training, owing to their capacity for instruction-following and structured reasoning. However, it remains unclear whether LLMs can reliably handle the multidimensional, theory-driven nature of sensemaking annotation - and whether performance is affected by definitional scaffolding, reasoning-oriented prompting, or enriched contextual representations of student knowledge.

\begin{figure}[ht!]
\centering
  \includegraphics[width=0.8\columnwidth]{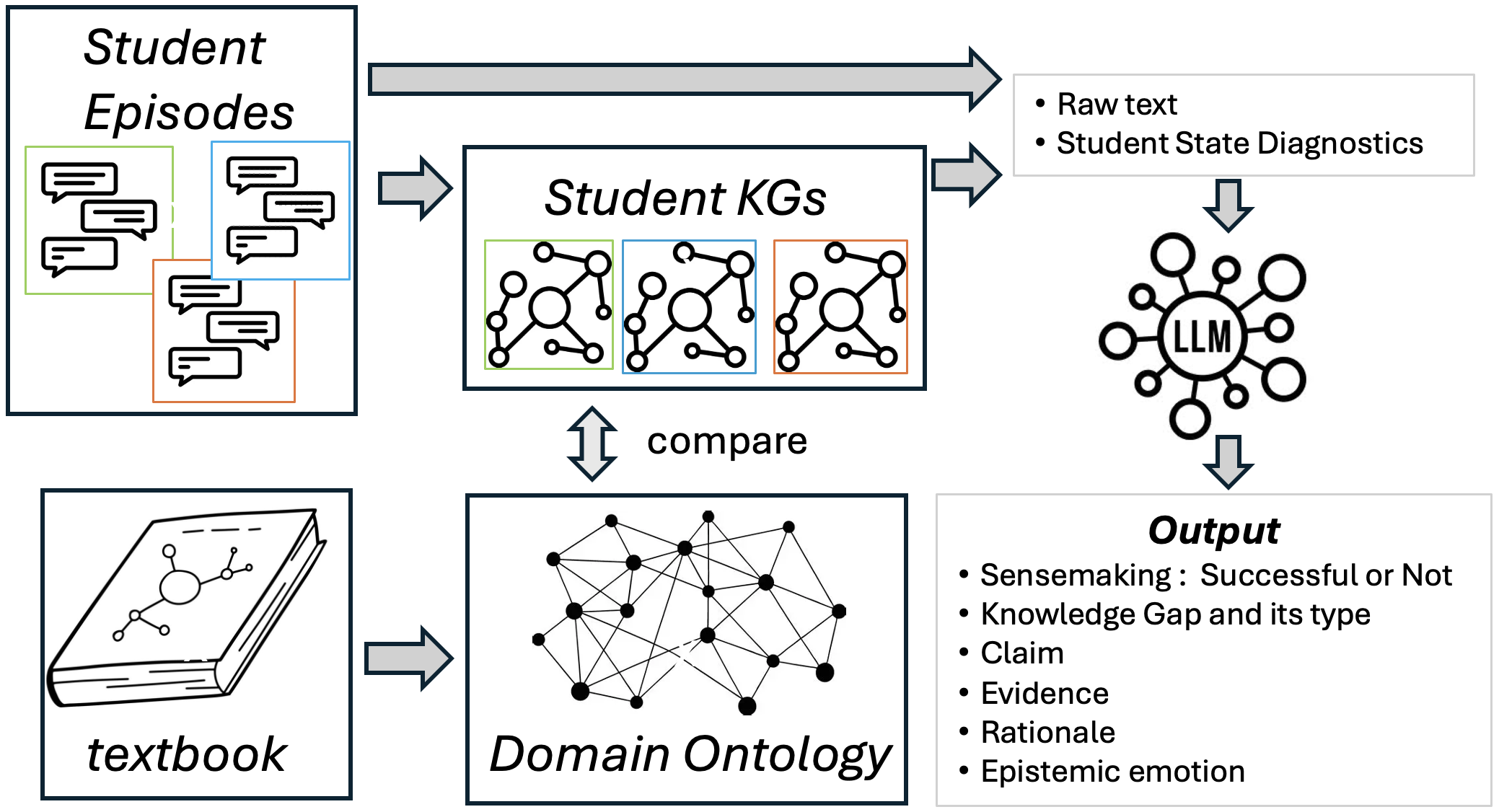}
  \caption{The proposed experimental pipeline for LLM-based sensemaking analysis. }
  \label{fig:pipeline}
\end{figure}

This study addresses these questions using expert-annotated collaborative dialogues from a Boyle's law laboratory activity (part of a first-year undergraduate chemistry curriculum). Using the framework presented in Figure~\ref{fig:pipeline}, we evaluate two mid-size instruct-LLMs across a range of prompting conditions - varying definitional context, reasoning mode (standard vs. chain-of-thought), and turn structure (single-turn vs. two-turn) - and assess whether augmenting dialogue with knowledge-graph (KG) derived knowledge-state diagnostics further improves sensemaking analysis.

The study is guided by the following three research questions: \textbf{RQ1.} How accurately can instruction-tuned LLMs identify successful and unsuccessful sensemaking, associated sensemaking features, and epistemic emotions from collaborative student dialogue? \textbf{RQ2. }How do definitional scaffolding, reasoning-enabled inference, and single-turn versus two-turn prompting affect model performance across these analytical tasks?
\textbf{RQ3.} Does augmenting student dialogue with KG-derived knowledge-state diagnostics improve LLM-based sensemaking analysis?

\textbf{Contributions.} From a learning sciences perspective, we provide a computational framework for examining not only whether sensemaking succeeds, but also how the knowledge expressed by learners is organized relative to the focal knowledge gap driving the episode. From a computational perspective, our joint evaluation of multiple model configurations on a rich, theory-driven annotation scheme yields systematic evidence of LLM instruction-following capabilities - including the effect of CoT prompting for a complex, multidimensional classification task.

\section{Related Work}
\paragraph{Sensemaking in learning.} \label{sec:relwork}
 Sensemaking can be defined as a dynamic process in which learners construct or revise explanations to resolve gaps or inconsistencies in their understanding by iteratively proposing, connecting, and evaluating ideas for coherence \cite{odden2019defining}. It is further characterized as an epistemic game comprising Step 0, organizing relevant prior knowledge; Step 1, recognizing a gap or inconsistency; Step 2, iterative explanation building; and Step 3, resolution of the focal knowledge gap \cite{odden2018sensemaking}.

The focal knowledge gap is therefore central to initiating and organizing sensemaking. It reflects a perceived uncertainty, inconsistency, or unresolved relation that becomes an epistemic problem for learners and may be conceptual, procedural, or epistemic. Such gaps can surface through recurring vexing questions (VQs), while explanation building can be characterized through claim-evidence-rationale (CER) structures, in which claims address the focal problem, evidence provides relevant observations or data, and rationale links evidence to claims through scientific principles or mechanisms \cite{haraldsrud2024using, hunter2021making, odden2019vexing, mcneill2006supporting}.

Sensemaking is neither necessarily linear nor inevitably successful. In successful sensemaking (SM), explanation building culminates in the resolution of the focal knowledge gap, whereas in unsuccessful sensemaking (USM), learners engage with the gap without reaching a coherent resolution \cite{gunes2026undergraduate}. Importantly, successful sensemaking does not require complete scientific correctness; success is determined by whether learners resolve the focal gap from their own perspective. This learner-centered criterion makes computational identification particularly challenging, as models must assess not only whether relevant scientific ideas are present, but whether those ideas become organized into an explanation that resolves the gap.

Epistemic emotions may provide complementary information about how learners experience unresolved problems during sensemaking. Emotions such as curiosity, confusion, frustration, and boredom are associated with learners’ appraisals of uncertainty and progress toward understanding \cite{pekrun2021self}. Consistent with dynamic accounts of affect during learning, these emotions may change as learners encounter cognitive impasses and attempt to resolve them; for example, confusion may accompany productive engagement when an impasse is addressed but may transition toward frustration or boredom when difficulties persist \cite{dmello2012dynamics,gunes2026undergraduate} . Accordingly, we treat emotional expressions as complementary indicators of learners’ experience rather than defining criteria of sensemaking itself. These components are interpreted jointly to determine whether a sensemaking episode is unfolding successfully or not.

\paragraph{LLMs for Modeling Learner Knowledge and Learning Processes.}
Modeling the collaborative sensemaking process is not a trivial task on several fronts: (1) it is a higher-order cognitive process that transcends topic and knowledge level: the goal is not to predict students' answers but to model the epistemic process itself; and (2) this process and its multidimensional indicators unfold over time, as real interactions are typically dialogue-based, open-ended, and rich in implicit signals that are not straightforward to capture. These challenges have led to growing interest in LLMs as a new paradigm capable of operating directly over conversational input \cite{wang2026large, cho2024systematic}. As shown by \citet{scarlatos2025exploring}, LLMs can infer student knowledge directly from dialogues, detecting misconceptions, tracking knowledge changes, estimating concept mastery, and predicting future performance. They also outperform some classical models in identifying uncertainty and partial understanding over time. However, \citet{scarlatos2025exploring} also show that purely prompt-based methods lacking structured knowledge integration remain prone to hallucination, and \citet{hooshyar2025problems} demonstrate that even extensive fine-tuning cannot fully overcome limitations in temporal coherence when assessing evolving knowledge states.

With LLMs, knowledge tracing can be performed through zero‑shot prompting, few‑shot in‑context learning, fine‑tuning, and integration of external structured knowledge sources \cite{neshaei2024towards, worden2026foundationalassist}. Advanced prompting strategies such as Chain‑of‑Thought (CoT) \cite{wei2022chain} and Chain‑of‑Pedagogy (CoP) \cite{lee2026rewarding} further support process-oriented reasoning by eliciting intermediate steps and teacher‑style guidance. Crucially, successful sensemaking does not require scientific correctness, only that the explanation resolves the learner’s gap. Therefore, while CoT improves models' reasoning capabilities, this does not entail that it improves understanding of sensemaking,  providing a challenging suite for understanding LLMs behavior. 

Beyond knowledge tracing, LLMs have shown strong potential for automating knowledge graph construction from domain text \cite{chen2025personalized, duan-etal-2026-rag}, though existing approaches lack domain-expert supervision, leaving them vulnerable to incompleteness and hallucinated relations - a risk particularly consequential in educational settings. Moreover, most ITS and LLM-based learner modeling approaches assume a single learner \cite{ma2014intelligent, vanlehn2011relative}, leaving collaborative epistemic processes largely unaddressed. By tracing knowledge, sensemaking features, and epistemic emotion as they co-evolve across a collaborative dialogue, our approach moves beyond outcome-focused evaluation toward process-sensitive analysis.

\paragraph{Emotion detection in NLP.}In collaborative sensemaking, epistemic emotions may provide complementary information about how learners experience and respond to unresolved knowledge gaps. Accordingly, emotion detection constitutes a complementary analytical dimension in our framework. NLP has developed substantial capacity for this task \cite{litman2004predicting, sun2025dialoguemllm}, with dedicated venues such as the Computational Affective Science workshop \cite{cas-2026-1} expanding the field well beyond early sentiment analysis toward affective computing, emotional language modeling, and emotion-aware dialogue systems. In this work, epistemic emotions serve as complementary temporal signals: shifts across successive dialogue segments (e.g., from curiosity to confusion) or from confusion to frustration  may indicate whether a sensemaking episode is progressing productively or heading toward an unsuccessful or interrupted outcome.

\section{Corpus} 
We use a corpus of collaborative dialogues collected from 18 first-year undergraduate chemistry education students during a Boyle’s Law laboratory activity \cite{gunes2026undergraduate}. Students worked in five small groups; further details of the instructional design and data collection are provided in the corpus paper. The corpus captures collaborative discourse around experimental design, observations, data interpretation, and explanation building. While 23 episodes (Table~\ref{tab:episode_stats}) is a modest number by standard NLP benchmarking norms, this corpus is not comparable to text classification datasets assembled from crowd-sourced or automatically labeled data. Each episode represents a lengthy, multi-turn collaborative dialogue (mean: 6500 words, 45 turns) annotated by domain experts against a theoretically grounded, multi-dimensional scheme, including sensemaking variations, sensemaking steps, focal knowledge gaps and their types, VQs, and CER structures. The per-episode annotation effort is therefore substantially higher than in standard sequence labeling or sentiment tasks. This places the present work closer to expert-annotated corpora in domains such as clinical NLP, legal reasoning, or scientific discourse analysis, where small but richly annotated datasets are often considered sufficient for evaluating model behavior on theoretically defined constructs.

 \begin{table}[h!]
\centering
\caption{Descriptive statistics for SM and USM episodes (for episode length and number of turns).}
\label{tab:episode_stats}
\resizebox{\columnwidth}{!}{%
\begin{tabular}{rccc}
\toprule
\multirow{2}{*}{\textbf{}} 
  & \textbf{Episode} 
  & \textbf{Mean Text Length} 
  & \textbf{Mean Turn Count} \\
  & \textbf{Count} 
  & \textbf{(SD)} 
  & \textbf{(SD)} \\\midrule
SM & 16 & 5863 words (3968) & 44 turns (39)  \\
USM & 7 & 8726 words (6010) & 46 turns (33) \\ \bottomrule

\end{tabular}%
}
\end{table}

\section{Sensemaking Modeling with LLMs}

Our experimental framework is illustrated in Figure~\ref{fig:pipeline} and proceeds in two stages. In the first stage, raw student dialogue episodes are provided directly to the model alongside definitional scaffolding, and we evaluate how accurately LLMs can identify sensemaking features from dialogue alone. In the second stage, we enrich the prompt with structured knowledge-state diagnostics derived from student knowledge state graphs (KGs), examining whether explicit representations of conceptual coverage and relational correctness improve model inference beyond what is available in surface-level text. Using 23 expert-annotated sensemaking episodes from a Boyle's Law laboratory activity, we evaluate two mid-size instruction-tuned LLMs across prompting conditions that vary definitional scaffolding, reasoning mode, and turn structure. For each episode, models are tasked with predicting several sensemaking dimensions jointly. Details of the prompt design and output schema are described below.

\subsection{Sensemaking Analysis From Raw Dialogue}
\paragraph{Models.} Two mid-sized instruction‑tuned models (Gemma‑3‑27B‑IT\footnote{\url{https://huggingface.co/google/gemma-3-27b-it}} and Qwen3‑32B\footnote{\url{https://huggingface.co/Qwen/Qwen3-32B}} ) are evaluated across four prompt variants that incrementally introduce definitional scaffolding for epistemic emotions and knowledge‑gap types, and across two inference modes: standard zero‑shot and reasoning‑enabled chain‑of‑thought. All configurations use greedy decoding for deterministic output, see App.~\ref{app:model_tech} for the runtime details. 

\paragraph{Prompt design.} Each dialogue episode is processed under four prompt variants (V1–V4) that incrementally augment the base instruction with definitional scaffolding. V1 includes only the core sensemaking framework; V2 adds epistemic‑emotion definitions; V3 adds knowledge‑gap type definitions; and V4 combines all three forms of scaffolding. The sensemaking framework follows the three-step model defined in Section~\ref{sec:relwork} (\textit{Step~1-3}), with unsuccessful sensemaking defined as progressing through at most a partial \textit{Step~2} \cite{odden2018sensemaking}. All prompt variants share an identical system persona (chemistry learning expert observing student groups), the same three-step sensemaking framework definition, and the same output instruction block. 

\paragraph{Task and output schema.} For each episode, the models are asked to return a single JSON object containing \textbf{seven} fields. The \textit{sensemaking} field indicates the sensemaking variation as either successful (SM) or unsuccessful (USM), based on the three-step sensemaking framework. Then it generates the focal knowledge gap driving the episode ($\leq$20 words), classifies the associated focal knowledge gap as conceptual, procedural, epistemic.  It also generates CER (claim-evidence-rationale) components of the episode ($\leq$20 words per each). Finally, it identifies the two most prominent epistemic emotions from a predefined set of eight: curiosity, surprise, confusion, anxiety, enjoyment, contentment, frustration, and boredom. 

Model outputs are compared against human-annotated ground-truth labels. The experimental design enables analysis of: (i)~the effect of definitional scaffolding on annotation accuracy; (ii)~the effect of CoT reasoning on each model; and
(iii)~cross-model differences in sensemaking, gap-type, and emotion classification performance.


\subsection{Integrating Knowledge‑State Diagnostics into LLM Inference}
\label{ssec:integratingKG}

While zeroshot prompting might be better suited for extracting the focal knowledge gap of the episode, VQ, or CER structures from raw student dialogues, as the theoretical definition of sensemaking describes,  sensemaking is a process of knowledge reorganization.  Here we leverage instruction‑tuned models to extract fine‑grained semantic structure from informal, multi-turn student conversations and transform it into machine‑interpretable representations that support downstream sensemaking modeling. We integrate knowledge-state diagnostics to explicitly represent how students organize their knowledge in relation to domain knowledge and the focal knowledge gap 
- information the model would otherwise have to infer from surface language alone. To our knowledge, no prior work has applied knowledge‑graph‑based diagnostics to sensemaking analysis.

\label{ssec:ontology} \paragraph{Subject-matter Knowledge Graph} As a reference model against which students' understanding can be measured, a structured representation of the target knowledge domain is required. Thus, we constructed a domain ontology automatically directly from curriculum text on \textit{Gases} Chapter containing \textit{Boyle's law}\footnote{A plain-text export of a chemistry textbook chapter (Chapter~10 :\textit{Gases} ) from libretexts \url{https://chem.libretexts.org/Bookshelves/General_Chemistry/}.}. Gemma-3-27B-IT is utilized to process each section of the text independently. A structured prompt written by domain expert (App.~\ref{fig:prompt_domain_ont_entity}) instructs the model to identify named entities across eight typed categories (\textit{Concept}, \textit{Law}, \textit{Equation}, \textit{Property}, \textit{Unit}, \textit{Scientist}, \textit{Device}, \textit{Process}) and return a one-sentence definition per entity as shown below. Entities are normalized to lowercase and merged by name; only the first occurrence is retained.

\begin{lstlisting}[language=json, breaklines=true, basicstyle=\small\ttfamily, , showstringspaces=false]
{ "name": "temperature",
  "type": "Property",
  "definition": "A measure of the average kinetic energy of molecules.",
  "section": "0" }
\end{lstlisting}
\vspace{-2mm}
A second LLM prompt (App.~\ref{fig:prompt_domain_ont_relation}) receives the full list of unique entity names and extracts typed relational triples using a fixed predicate vocabulary also selected by domain expert such as \texttt{is\_subconcept\_of}, \texttt{measured\_by}. The full list is provided in the prompt template. Entities are instantiated as OWL individuals under a typed class hierarchy using \texttt{owlready2}. Object and data properties are populated with extracted relations and definitions. OWL reasoner (HermiT) check unsatisfiable classes, domain or range violations, and cycles in transitive properties. The final subject-matter graph on Boyle's Law contains 112 unique entities and 609 triples. 

\paragraph{Students' Knowledge State Construction}
\label{ssec:studentKG}

Assessing conceptual understanding from classroom dialogue requires mapping informal, multi-turn student utterances onto formal domain vocabulary, identifying which concepts and relations students express and whether they do so correctly, and evaluating how their current knowledge state relates to the focal question. A per-episode knowledge graph enables structured, scalable comparison against the domain reference. 
Each episode is processed in two sequential LLM calls for \textit{entity spotting} and \textit{relation extraction}. Entity names from the domain KG are extracted as a controlled vocabulary. The same relation types are carried forward. The model is given the full controlled vocabulary and the raw episode text. It identifies which domain entities students mention, reference and annotates each match with a
short rationale and label match (\textit{correct}, \textit{incorrect}, or \textit{partial}). For the entities identified, the model extracts relational triples (subject, predicate, object) that students explicitly or implicitly express.  Hallucinated entities (names absent from the vocabulary) and invalid predicates (outside the allowed relation set) are flagged and discarded. Then, each episode's validated entities and triples are converted into a \texttt{NetworkX} directed graph (\texttt{DiGraph}). Node attributes include entity type, definition (pulled from the domain KG), how the entity was mentioned, and label match. The focal question of the episode (annotated by the domain expert) is also converted to KG using the same method. 

\paragraph{Knowledge State Diagnostics}

The knowledge-state diagnostic is obtained by comparing each student KG against both the domain KG and the focal question KG. We compute a rich set of diagnostic metrics, including node recall, edge recall, cosine similarity between the student and domain graphs, the number of missing domain concepts, and the number of student-expressed relations that contradict the domain ontology. The same comparison is performed against the focal question KG to assess topical relevance. Additionally, structural similarity between the student KG and both reference graphs is computed using node and edge counts as feature vectors. These diagnostics, together with their expert-curated interpretation guidelines, are provided to the LLM as additional context.

Prior to this, a domain expert conducted a quantitative validation analysis of the student knowledge-state graphs and their diagnostics for half of the episodes. Successful and unsuccessful sensemaking were characterized by qualitatively distinct patterns of knowledge organization. In SM, activated concepts became organized into a relational structure relevant to the focal knowledge gap and sufficient to support its resolution — not necessarily error-free, but coherent enough to close the gap. In USM, activated knowledge failed to develop into a gap-resolving explanatory structure, whether through fragmentation, contradictory relations, or the activation of concepts that remained irrelevant to the focal gap. The distinction thus lies not in the number or correctness of concepts alone, but in whether they become integrated into a structure that resolves the driving knowledge gap.

This contrast is illustrated by two representative episodes (App. Figure~\ref{fig:studentKGs_samples}). In SM5, students were initially uncertain about how to coordinate water-column height, hydrostatic pressure, atmospheric pressure, and mmHg conversion when applying P\textsubscript{gas} = P\textsubscript{h} + P\textsubscript{atm}. As the discussion progressed, these resources became organized around the focal procedural gap, culminating in explicit resolution and expressed \textit{contentment} and \textit{enjoyment}. In USM4, students identified an apparent inconsistency between their data and Boyle's Law - interpreting both pressure and volume as increasing - and activated many relevant concepts, yet these never stabilized into a coherent explanatory structure. The group cycled repeatedly back to the unresolved question, ending in sustained confusion and abandonment. Overall, the qualitative expert validation confirmed that the student KGs and their derived diagnostics were consistent with the sensemaking variation of each episode. Importantly, these diagnostic patterns were also found to be consistent with the raw episode content, suggesting that the KG-derived representations capture meaningful epistemic signal beyond surface-level dialogue features.
 
\section{Results}
Given the sequential and multi‑dimensional nature of sensemaking identification, we evaluate several zero‑shot prompting configurations. Specifically, we vary (i) whether explicit reasoning is enabled, (ii) whether the task is handled in a single turn or decomposed into a two‑turn setup, and finally (iii) whether high‑level knowledge state signals are injected into the prompt. These conditions allow us to examine how different prompting strategies shape model performance across the sensemaking tasks.

\subsection{Base models with Raw Dialogs} 

We compare Gemma-3-27B and Qwen3-32B across two reasoning/thinking settings (Disabled and Enabled) on identifying sensemaking features either with single-turn or two-turn prompting. All configurations are fed the same 23-episode corpus with a ground-truth sensemaking binary label (1 = successful, 0 = unsuccessful).
The class distribution is imbalanced (successful (16/23)), making a majority-class baseline accuracy 69.6\%.

\paragraph{Single-turn prompting.} As shown in Table~\ref{tab:basemodels}, thinking mode benefits both models, particularly for USM episodes. Definitional scaffolding shows mixed effects (App. Table~\ref{tab:basemodels_prompts}): while Gemma remains consistent across all 4 prompt variants, adding epistemic emotion definitions to the thinking prompt (V2) hurts QWEN3 performance. The 3 hardest episodes  are all unsuccessful (USM), sharing dialogue features that superficially resemble successful sensemaking (e.g., partial explanations models treat as resolved).


\begin{table}[h]
\centering
\resizebox{\columnwidth}{!}{%
\begin{tabular}{rcclll}
\toprule
\textbf{Model Config} & \textbf{Best Prompt} & \textbf{F1-SM}& \textbf{F1-USM} & \textbf{Cohen $\kappa$} \\ \midrule
Gemma-3-27B No-Thinking & V2 & 81\% & 40\% & 0.22 (weak) \\
Gemma-3-27B Thinking & V1 & 84.9\% & \textbf{61.5}\%& 0.45 (moderate) \\ \midrule
Qwen3-32B No-Thinking & V1-3 & 86.5\% & 44.4\%& 0.36 (fair) \\
\textbf{Qwen3-32B Thinking} & \textbf{V3} & \textbf{88.9\%} & 60\%& \textbf{0.51 (moderate)} \\ \midrule
\end{tabular}%
} 
\caption{Sensemaking Classification (single-turn)}
\label{tab:basemodels}
\end{table}

 Knowledge gap type prediction was largely unsuccessful, with models defaulting to "conceptual" across nearly all instances. Thinking mode partially mitigated this overgeneralization; in particular, the thinking model with knowledge gap definitions improved classification accuracy for "procedural" gaps to approximately 30\% (see App. Table~\ref{tab:KGT_results}).
Based on overall sensemaking performance, we proceed with Qwen3 for the remaining analyses, as it achieves the highest agreement with human annotations ($\kappa = 0.51$).

\paragraph{Two-turn prompting.} We also experiment with an two-turn design (see App.~\ref{app:twoturn}). Turn 1 elicits a free-text step-by-step analysis grounded in the sensemaking framework, and Turn 2 presents that analysis back to the model and requests only the structured JSON output, thereby decoupling the reasoning process from the output formatting constraint and preventing the schema from collapsing the inference into surface-level pattern matching \cite{zhang2025survey, han-etal-2025-language}. 

Upper part of Table~\ref{tab:CoTresults} shows the effect of turn structure across two inference modes. While two-turn hurts the classification of SMs, it improves the more challenging unsuccessful case (USM) identification. Yet, on average no effect on the alignment to human labels is observed. But when we look at the instance level,  \textit{two-turn} (V2) uniquely solves two USM cases (missed by all others) and \textit{Thinking} (V3) uniquely solves another USM (missed by all others). These results indicate that the two configurations capture different aspects of the task.

One caveat of the two-turn approach is that while it improves USM detection, it also hurts performance on other sensemaking indicators, such as knowledge gap type and epistemic emotion classification (as discussed later). This observation raises an important question for future research: whether models make correct decisions for the right reasons, and how prompt configuration interacts with a model's reasoning process. Such interpretability is critical if LLMs are to be reliably deployed as analytical tools in educational settings.

\begin{table}[]
\centering
\resizebox{\columnwidth}{!}{%
\begin{tabular}{rcclll}
\toprule
\textbf{Model Config} & \textbf{Best Prompt} & \textbf{F1-SM}& \textbf{F1-USM} & \textbf{Cohen $\kappa$} \\ \midrule
No-Thinking, single-turn & V1, 3, 4 & 86.5\% & 44.4\%& 0.36 (fair) \\
No-Thinking, two-turn & V2 & 75.9\% & 58.8\%&  0.36 (fair) \\ 
Thinking, single-turn  & V3 & \textbf{88.9\%} & 60.\%& 0.51 (moderate)\\ 

Thinking, two-turn  & V3 & 76.9\% & 70.\%& 0.50 (moderate) \\ \midrule
\textbf{KG-informed Variants } & & & & \\
No-Thinking, single-turn  & - & 86.5\% & 44.\%& 0.36 (fair) \\
No-Thinking, two-turn  & - & 76.9\% & 70.0\%& 0.50 (moderate)\\
Thinking, single-turn  & - & 87.5\% & \textbf{71.4}\% & \textbf{0.60 (moderate)}  \\
Thinking, two-turn  & - & 66.7\% & 63.6\% &  0.38 (fair)\\
\bottomrule
\end{tabular}%
} 
\caption{Sensemaking Classification (Qwen3-32B)}
\label{tab:CoTresults}
\end{table}

\subsection{Effect of Knowledge Diagnostics}
\label{ssec:kg_results}
The previous classification experiments rely exclusively on raw dialogue text. An open question is whether grounding the model's analysis in the quantitative structural evidence produced by the student KG pipeline improves classification accuracy. We hypothesize that injecting per-episode KG comparison metrics as additional prompt context allows the model to reason over objective indicators of conceptual coverage and relational correctness, rather than inferring these solely from surface language. Using Qwen-3-32B, we ask model to output the same annotation scheme as before. In addition to the baseline prompt with sense-making framework, we insert a block to the prompt which contains the Student KG diagnostics and guideline on how to interpret it, as shown in App. Figure~\ref{fig:prompt_llm_classification_part2}.

Table~\ref{tab:CoTresults} (lower panel) presents results for the KG-informed configurations. Without thinking mode, incorporating KG diagnostics introduced noise rather than useful signal, slightly degrading sensemaking classification. Similarly, combining KG diagnostics with two-turn thinking hurt performance substantially, suggesting that the model's internal reasoning trace is considerably more effective than the explicit diagnostic grounding in this configuration. By contrast, KG diagnostics provided meaningful grounding for the thinking model in the single-turn setting: SM performance remained stable relative to the no-KG baseline, while USM detection improved substantially, yielding the highest alignment with human annotations across all tested configurations (
$\kappa = .60$). This suggests that the quantitative KG evidence appears to provide the crucial signal that overcomes the model's SM bias. Yet, USM is the hardest case; this is where KG integration earns its value.  For knowledge gap type classification, KG diagnostics produced only marginal improvement; even the best-performing configuration — thinking mode with two-turn prompting - reached only 50\% accuracy on the "conceptual" gap type (App. Table~\ref{tab:KGT_results}), indicating that this remains an open challenge.

\begin{figure}[ht!]
\centering
  \includegraphics[width=.75\columnwidth]{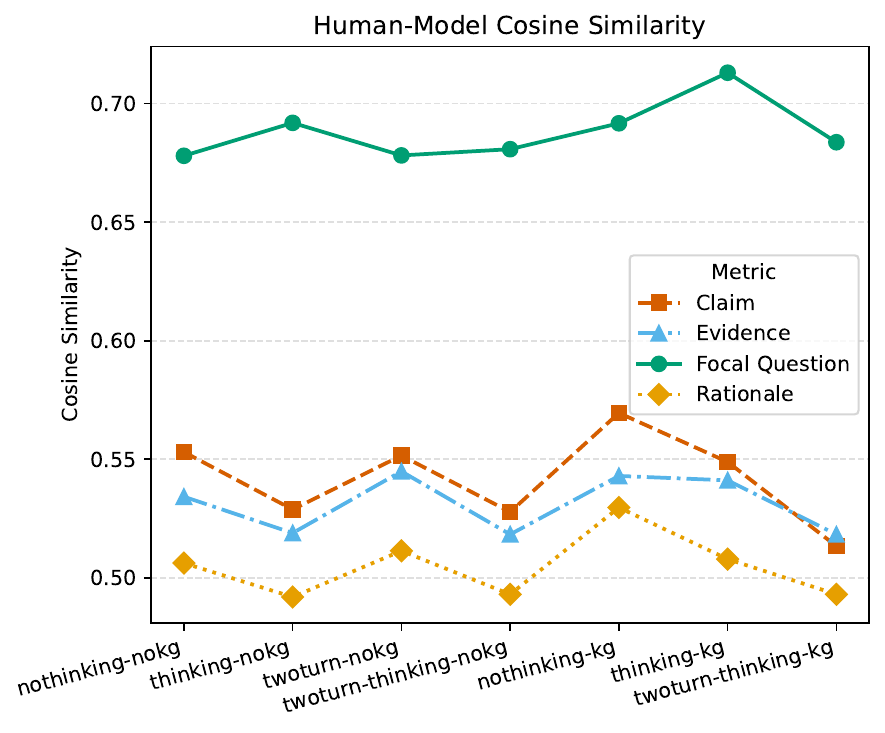}
  \caption{Cosine Similarity between expert-produced and model-generated features.}
  \label{fig:similarity_plot}
\end{figure}

\textbf{Human-model alignment at focal questions and CER components.}
\label{ssec:cer_results}
Figure~\ref{fig:similarity_plot} compares cosine similarities, calculated using pretrained model (SciBERT\footnote{\url{https://huggingface.co/allenai/scibert_scivocab_uncasedusing}}, \cite{beltagy-etal-2019-scibert}  between expert-produced and model-generated focal questions and CER components across seven Qwen3-32B configurations varying in thinking mode, turn structure, and the inclusion of KG-derived knowledge-state diagnostics. Overall, the simpler no-thinking configurations showed comparatively strong human-model alignment, particularly for claim and rationale. Enabling thinking did not consistently improve similarity, and neither two-turn prompting nor the inclusion of KG-derived diagnostics produced a uniform advantage across the four components. Instead, the effects of these prompting modifications varied by component and configuration. Focal question identification showed the highest and most stable similarity across configurations, whereas similarity for the CER components was generally lower. Among the CER components, rationale consistently showed the lowest similarity to expert annotations, indicating that identifying the reasoning linking evidence to a claim was the most challenging aspect of the extraction task. 
These results show that, unlike SM/USM identification, different configurations benefit different features: for example, \textit{nothinking‑KG} for Claim and Rationale, \textit{two‑turn} for Evidence, and \textit{thinking‑KG} for the focal question. Together, these patterns highlight the complexity of the problem and the difficulty of achieving a single configuration that performs well across all features. 

\textbf{Epistemic Emotion Prediction.} 
\label{ssec:emotion_results}
Epistemic emotions are another indicator of how sensemaking unfolds. Since knowledge diagnostics were not directly designed to influence emotion classification, we analyze the distribution of emotions predicted for SM and USM episodes across the four Qwen3-32B configurations without diagnostics integration (App. Figure~\ref{fig:emotion_pies}). Across configurations, curiosity, confusion, contentment, and frustration accounted for the majority of model predictions, whereas enjoyment and anxiety were rarely identified, and surprise and boredom were entirely absent. The emotional profiles of SM and USM episodes showed some differentiation, although the magnitude of this contrast varied substantially across prompting configurations. In the no-thinking condition, curiosity was the most frequently predicted emotion for both SM (43\%) and USM (45\%) episodes. SM predictions were additionally characterized by contentment (31\%) and confusion (25\%), whereas USM predictions included a higher proportion of frustration (21\%) and a lower proportion of contentment (16\%). A similar pattern was observed in the thinking condition: SM episodes were dominated by curiosity (41\%), confusion (29\%), and contentment (26\%), whereas frustration increased to 23\% for USM episodes, although curiosity remained the most frequent prediction (38\%). 

The distinction became more pronounced in the two-turn configurations. Yet, while single-turn configs generate emotion predictions for each episode; two-turn configs drop to 25–43\% because the JSON template's emotion field is often omitted or null when the model focuses on the analytical reasoning. Across the two-turn configurations, USM episodes were more strongly characterized by the combination of confusion and frustration, whereas SM episodes retained comparatively greater proportions of curiosity and contentment. Overall, these results suggest that the model captured some differences in the emotional profiles associated with SM and USM, but that these differences were strongly dependent on the prompting configuration. 

Notably, confusion was prominent in model predictions for both SM and USM episodes, suggesting that confusion is not associated with certain sensemaking outcome. This pattern is conceptually consistent with our previous analysis of the same learning context, in which confusion occurred in both task-level successful and unsuccessful sensemaking but was embedded in different emotional contexts \cite{gunes2026undergraduate}. In the successful one, confusion co-occurred with positive activating emotions and was resolved through collaborative reasoning, whereas in the unsuccessful one it occurred alongside negative emotions such as frustration and boredom. The present LLM results show a related distinction at the aggregate level: particularly in the two-turn configurations, USM predictions were characterized not simply by greater confusion but by its co-occurrence with substantially greater frustration and reduced contentment. This suggests that distinguishing SM from USM may require attention to configurations of epistemic emotions rather than individual emotion labels in isolation.

\section{Conclusion}

This study examined whether instruction-tuned LLMs can support multidimensional analysis of collaborative sensemaking and whether KG-derived diagnostics improve inference beyond raw dialogue. Reasoning-enabled prompting improved USM identification (\textbf{RQ1}), and KG-state diagnostics provided additional grounding when combined with the appropriate reasoning configuration (\textbf{RQ3}) - yielding the highest human-model alignment ($\kappa = .60$). However, performance remained uneven across knowledge gap types, CER components, and epistemic emotions, and two-turn prompting produced mixed effects across dimensions (\textbf{RQ2}). This configuration-dependence suggests that sensemaking analysis may ultimately require ensemble or adaptive prompting strategies tailored to specific dimensions. These findings also indicate that LLM-based sensemaking analysis does not yet operate as a standalone annotation tool, but supports an expert-in-the-loop workflow in which models surface theoretically grounded features for human interpretation. More broadly, the study demonstrates the value of representing not only what concepts students mention, but how those concepts are organized in relation to the focal knowledge gap. Moreover, this theory-driven analysis enhances our understanding of LLM capabilities, the interactions among various configuration choices, and the inherently multidimensional nature of the sensemaking task. Future work will extend this framework temporally by tracking changes in student KGs, CER structures, and epistemic emotions as sensemaking unfolds. 

\section*{Limitations}

The corpus comprises 23 episodes from a single chemistry topic. Although small by NLP benchmarking standards, the corpus reflects the practical realities of expert-annotated educational corpora: the annotation scheme is theoretically grounded, multi-dimensional, and requires substantial domain expertise to apply reliably. Our aim is not to establish broad statistical generalization but to evaluate whether LLMs can operationalize a complex theoretical construct - a question that is meaningful even at this scale. Nonetheless, replication on larger and more topically diverse corpora remains an important direction for future work.

We also restrict our experiments to two mid-sized, open-weight LLMs. Our consortium prioritizes sustainable and accessible AI, making mid‑size models a practical choice for research and deployment. Rather than conducting a broad model benchmark, our aim is to examine how LLM-derived knowledge-state diagnostics and different prompting configurations interact with the multidimensional components of sensemaking. Consequently, the reported findings should not be interpreted as representative of the performance of smaller, larger or proprietary models.

\section*{Ethical Section}
The study uses an existing corpus collected within a nationally funded research project. Data collection was conducted with prior approval from the relevant institutional ethics committee and with informed consent from all participants. The present analyses use anonymized student dialogue data and expert annotations; no personally identifiable information is included. Data are used in accordance with the scope of the original research project and applicable institutional ethical requirements.



\bibliography{custom}

\appendix
\section{Appendix}
\label{sec:appendix}

\subsection{Sample Expert Annotation}
\label{sec:episode_sample}

\begin{figure}[h!]
  \includegraphics[width=0.9\columnwidth]{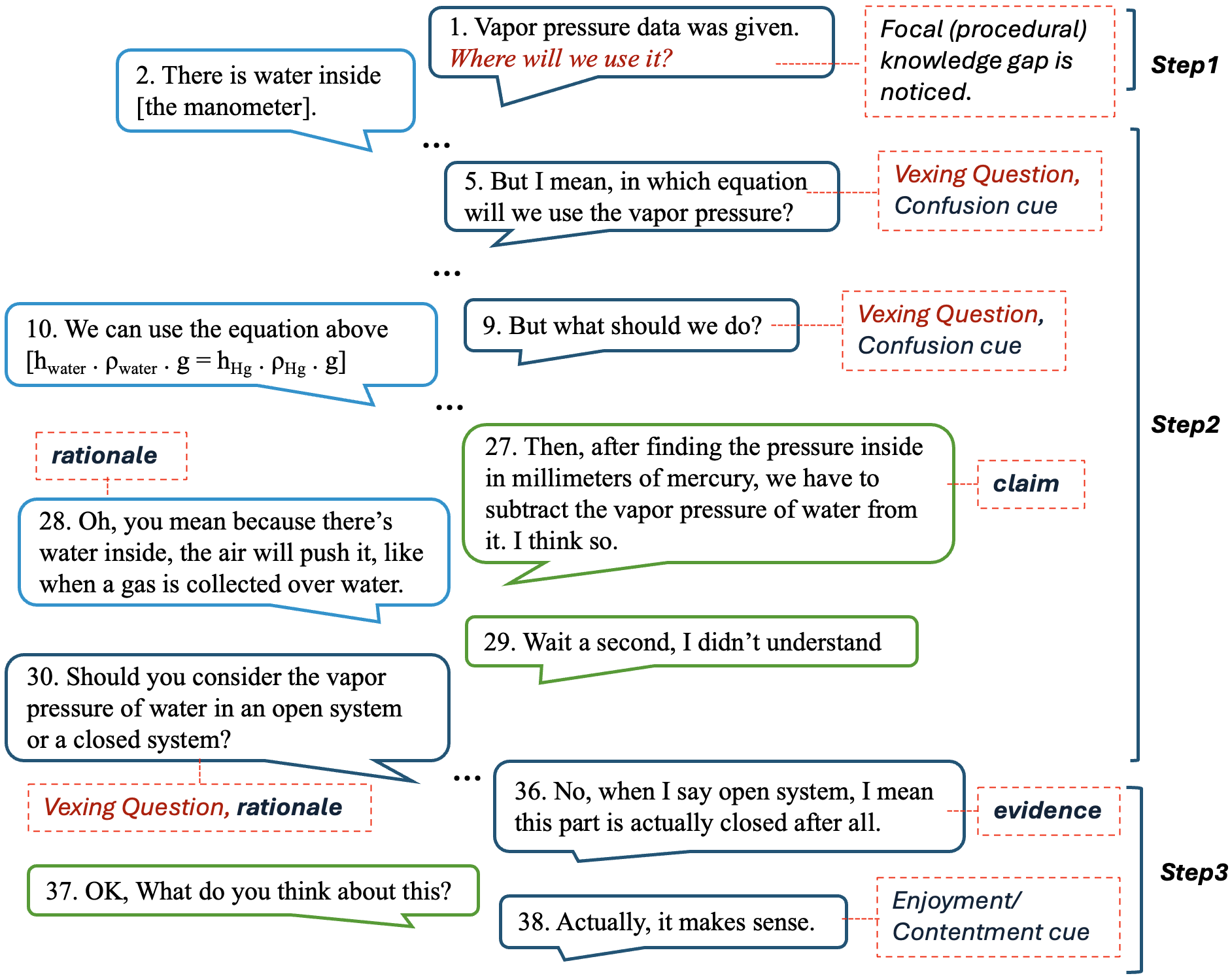}
  \caption{Excerpts from a successful collaborative sensemaking episode illustrating the temporal unfolding of sensemaking across Steps 1–3, the focal knowledge gap, recurring VQs, CER components, and complementary emotional cues.}
  \label{fig:sample_dialogue}
\end{figure}

Figure~\ref{fig:sample_dialogue} illustrates how sensemaking features can emerge and develop over time within a successful collaborative sensemaking episode.

\subsection{Computational Environment and Runtime Details}
\label{app:model_tech}

All experiments were run on a single NVIDIA A100 (40\,GB HBM2) GPU via Google Colab Pro, with approximately 84\,GB of system RAM available. Both models are loaded in 4-bit NF4 quantisation with double quantisation enabled (\texttt{bnb\_4bit\_use\_double\_quant=True}) via \texttt{bitsandbytes}. All inference uses greedy decoding with a 2048-token generation budget.

\begin{itemize}
    \item \textbf{Gemma-3-27B-IT} ($\approx$14\,GB VRAM): evaluated under thinking-off and thinking-on conditions. Because Gemma-3-IT does not expose a native thinking flag, the thinking-on condition is emulated through an explicit \texttt{<reasoning>}~\ldots~\texttt{</reasoning>} prompt prefix that elicits step-by-step reasoning before the final answer.

    \item \textbf{Qwen3-32B} ($\approx$18\,GB VRAM): evaluated with \texttt{enable\_thinking=False} (thinking-off) and \texttt{enable\_thinking=True} (thinking-on). In the thinking-on condition, the model produces a \texttt{<think>}~\ldots~\texttt{</think>} internal reasoning trace prior to the final JSON-formatted response.
\end{itemize}

For Qwen3, instructions are delivered via a \texttt{system}/\texttt{user} message split. For Gemma-3, which does not support a native system role, all content is concatenated into a single \texttt{user} turn. All prompt variants for a given episode are run sequentially under the same loaded model configuration.

\subsection{Two-turn Prompt Design}
\label{app:twoturn}
When a model is asked to produce a JSON object directly, the generation task has two competing demands simultaneously - think through the evidence and conform to a rigid schema. In practice, the schema wins. The model anchors on the output format early in the generation, which means the classification fields are filled by pattern-matching rather than by genuine chain-of-thought inference. This is especially problematic for nuanced multi-step frameworks like sensemaking classification, where the correct label depends on a sequential causal chain (gap → explanation attempt → resolution) that cannot be recovered from surface-level text features alone. In Turn 1, the model receives the framework definitions and the episode, and is instructed to reason in plain text: identify evidence for each step, evaluate explanation quality, and reach a judgment. No output schema is imposed. The output is stored as reasoning trace. In Turn 2,  the entire conversation history including the Turn 1 analysis is fed back as context, and the model is asked only to convert its own prior reasoning into the JSON output schema. Because the reasoning is already done and is part of the prompt context, the model is not generating the reasoning and the label simultaneously. The label is a transcription of a conclusion already reached, not a first-pass pattern match.

\subsection{Epistemic Emotion Distributions}

\begin{figure*}[h!]
\centering

\subfloat[Without Knowledge Diagnostics\label{fig:emot}]
{\includegraphics[width=0.4\textwidth]{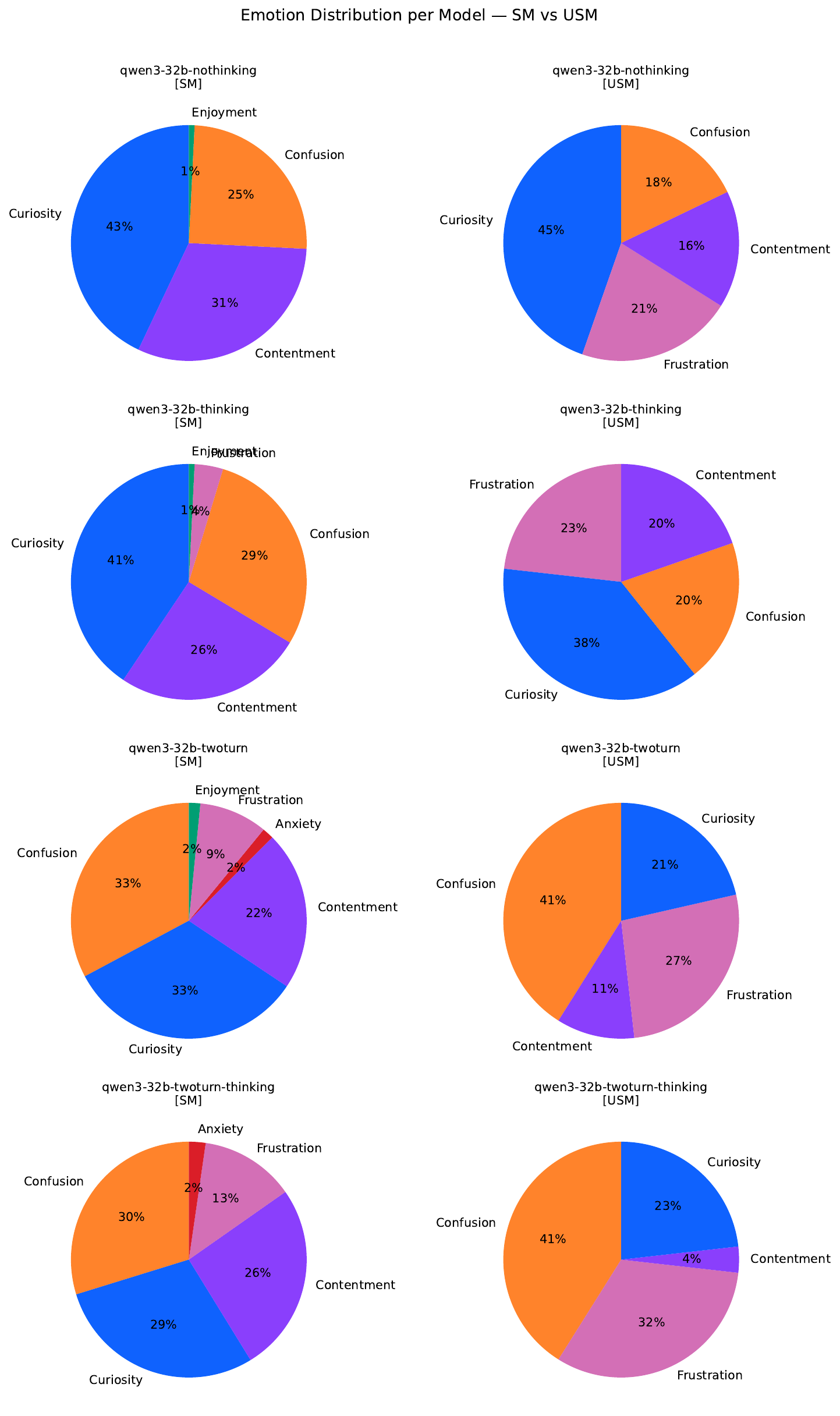}}
~
\subfloat[With Knowledge Diagnostics\label{fig:emot_kg}]
{\includegraphics[width=0.4\textwidth]{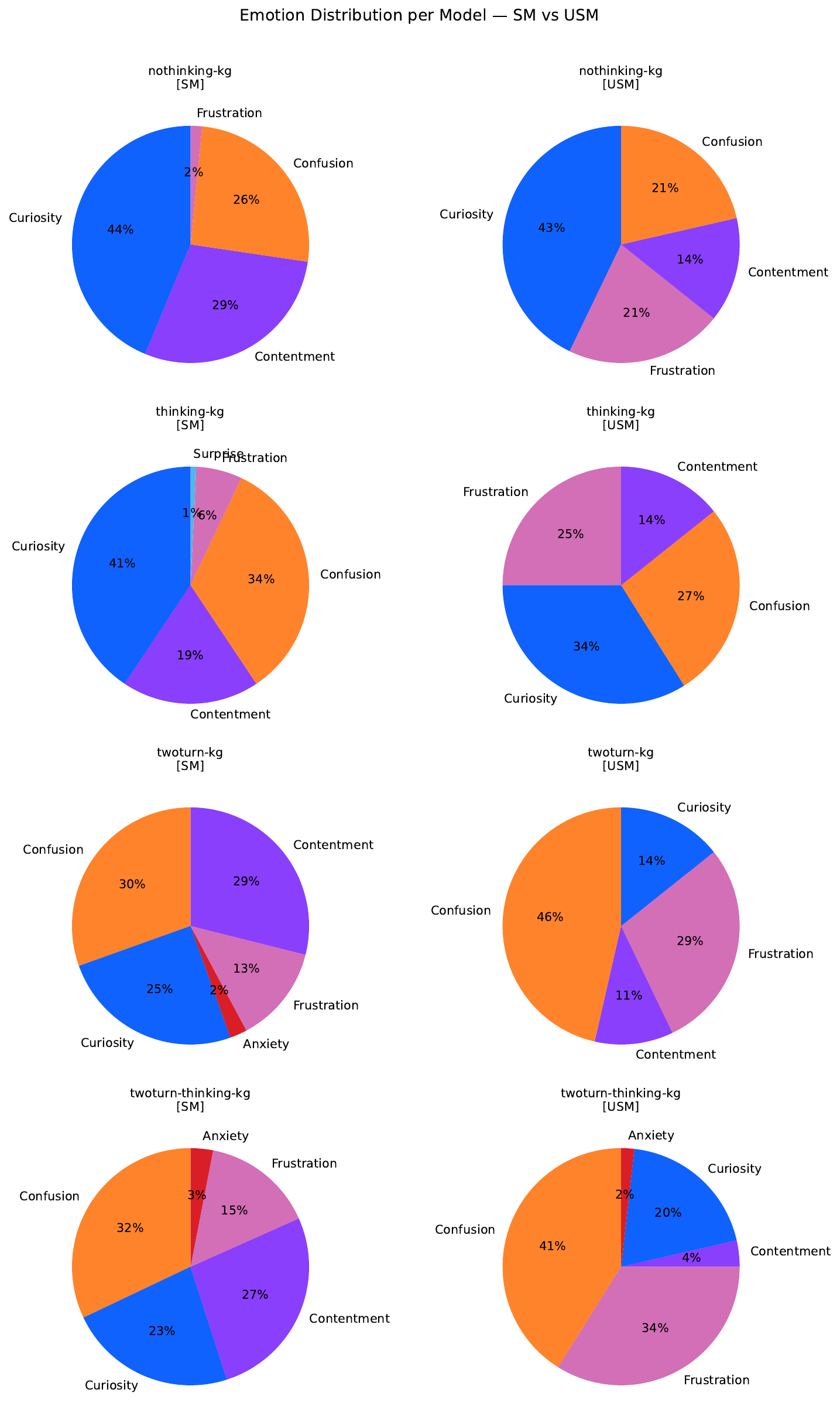}}

\vspace{-1mm}	
\caption{QWEN3-32B Models' Emotion Distribution for SM~\&~USM}
\label{fig:emotion_pies}
\vspace*{-3mm}
\end{figure*}

Figure~\ref{fig:emotion_pies} the distribution of epistemic emotions predicted for SM and USM episodes across all Qwen3-32B configurations with and without diagnostics integration.

\begin{figure*}[t]
\centering

\begin{tcolorbox}[
    title={Entity Extraction},
    fonttitle=\bfseries,
    coltitle=white,
    width=\textwidth,
    breakable,
    colback=white,
    colframe=black
]

\small\ttfamily\obeylines

\#\#\# ENTITY\_PROMPT = \\
You are a chemistry education expert building a formal OWL ontology for a textbook chapter on Gases. \\

From the provided text, extract ALL specific, named entities that are central to the chapter's learning objectives. \\
Classify each entity into EXACTLY ONE of these categories: \\

- Concept       : A specific chemistry principle, phenomenon, or theoretical idea (e.g., "compressibility", "intermolecular forces", "absolute zero", "mole fraction", "mean free path") 
- Law           : A named empirical or theoretical scientific law (e.g., "Boyle's Law", "Dalton's Law of Partial Pressures", "Graham's Law") 
- Equation      : A named mathematical relationship or formula (e.g., "Ideal Gas Law", "van der Waals Equation", "PV = nRT") 
- Property      : A measurable physical or chemical attribute (e.g., "pressure", "volume", "temperature", "density", "molar mass", "kinetic energy", "rms speed") 
- Unit          : A standard unit of measurement (e.g., "pascal", "atmosphere", "kelvin", "liter", "torr", "J/(Kmol)") 
- Scientist     : A person credited with a discovery, law, or theory (e.g., "Robert Boyle", "Johannes van der Waals", "Thomas Graham", "Evangelista Torricelli") 
- Device        : A physical instrument or apparatus used for measurement or demonstration (e.g., "barometer", "manometer", "J-shaped tube") 
- Process       : A physical or chemical change or mechanism (e.g., "diffusion", "effusion", "liquefaction", "phase transition", "condensation") \\

\vspace{2ex}
\#\#\#  RULES:  \\
1. Extract ONLY entities explicitly mentioned or directly defined in the text.  
2. Do NOT extract overly generic terms (e.g., "gas", "matter", "substance", "energy") unless they are part of a specific named concept.  
3. Definitions must be concise (max 15 words) and scientifically accurate.  
4. Return ONLY valid JSON. No markdown fences, no extra text.  

\vspace{2ex}
\#\#\# Output format: 
[ \{ "name": "exact entity name", 
    "type": "Concept | Law | Equation | Property | Unit | Scientist | Device | Process", 
    "definition": "concise definition", 
    "section": "\{section\_id\}" \}]\\

    \end{tcolorbox}

\caption{Prompt Template 1a: Domain Ontology Construction (Entity Extraction)}
\label{fig:prompt_domain_ont_entity}
\end{figure*}

\begin{figure*}[t]
\centering

\begin{tcolorbox}[
    title={Relation Extraction},
    fonttitle=\bfseries,
    coltitle=white,
    width=\textwidth,
    breakable,
    colback=white,
    colframe=black
]
\small\ttfamily\obeylines

Identify the most important relationships between them.\\
Use only these relation types:\\
- is\_subconcept\_of          (e.g., "Boyle's Law" is\_subconcept\_of "Gas Laws")
- derived\_from              (e.g., "Ideal Gas Law" derived\_from "Boyle's Law")
- is\_special\_case\_of        (e.g., "Boyle's Law" is\_special\_case\_of "Ideal Gas Law") 
- directly\_proportional\_to  (e.g., "gas volume" directly\_proportional\_to "temperature") 
- inversely\_proportional\_to (e.g., "gas volume" inversely\_proportional\_to "pressure") 
- depends\_on                (e.g., "gas density" depends\_on "molar mass") 
- formalized\_by             (e.g., "pressure-volume relationship" formalized\_by "Boyle's Law") 
- held\_constant\_in          (e.g., "temperature" held\_constant\_in "Boyle's Law") 
- measured\_by               (e.g., "pressure" measured\_by "barometer") 
- has\_unit                  (e.g., "pressure" has\_unit "pascal") 
- discovered\_by             (e.g., "Dalton's Law" discovered\_by "John Dalton") 
- describes                 (e.g., "Kinetic Molecular Theory" describes "gas behavior")
- applies\_to                (e.g., "van der Waals Equation" applies\_to "real gases") 
- is\_correction\_of          (e.g., "van der Waals Equation" is\_correction\_of "Ideal Gas Law") 
- is\_explained\_by           (e.g., "Boyle's Law" is\_explained\_by "Kinetic Molecular Theory") 
- explains                  (e.g., "molecular collisions with container walls" explains "gas pressure") 
- results\_in                (e.g., "increasing temperature" results\_in "increased molecular speed") 
- has\_constant              (e.g., "Ideal Gas Law" has\_constant "gas constant R") 

\vspace{2ex}
\#\#\# RULES:  \\
1. Use ONLY entity names from the provided list. Do not invent new entities.  
2. Focus on relationships explicitly stated or strongly implied in the textbook. 
3. Maintain correct directionality: subject -> predicate -> object.  
4. Return at most 60 of the most meaningful triples.  
5. Return ONLY valid JSON. No markdown fences, no extra text.  

\vspace{2ex}
\#\#\# Output format:  
[ \{ "subject": "entity A", "predicate": "relation\_type", "object": "entity B" \} ]\\

\end{tcolorbox}
\caption{Prompt Template 1b: Domain Ontology Construction (Relation Extraction)}
\label{fig:prompt_domain_ont_relation}
\end{figure*}

\begin{figure*}[t]
\centering
\begin{tcolorbox}[
    title={LLM Classification from raw text (part 1)},
    fonttitle=\bfseries,
    coltitle=white,
    width=\textwidth,
    breakable,
    colback=white,
    colframe=black
]
\small\ttfamily\obeylines

\vspace{2ex}
\#\#\# system\_role :
You are an expert in analyzing chemistry learning conversations.
You observe groups of students working on \"Boyle's Law\" experiments.

\vspace{2ex}
\#\#\# Knowledge gap types:
Conceptual gaps refer to difficulties in explaining phenomena due to incomplete, conflicting, or insufficient conceptual understanding.
Procedural gaps encompass students' difficulties understanding and applying scientific methods, problem-solving processes, or experimental procedures.
Epistemic gaps involve uncertainty about the nature and justification of scientific knowledge.

\vspace{2ex}
\#\#\# emotion\_lists:  'Curiosity, Surprise, Confusion, Anxiety, Enjoyment, Contentment, Frustration, Boredom'

\vspace{2ex}
\#\#\# emotion\_definitions 
- Curiosity: A desire to learn, fill a knowledge gap, resolve an inconsistency, or understand a phenomenon.
- Surprise: An immediate reaction to unexpected information, data, or observations that conflict with prior knowledge or expectations.
- Confusion: A state of cognitive disequilibrium that occurs when students are unsure how to fill a knowledge gap or resolve an inconsistency.
- Anxiety: Tension or worry that arises when students perceive the knowledge gap or inconsistency as difficult to resolve or beyond their current understanding.
- Enjoyment: A positive emotion experienced while exploring, understanding, or successfully explaining a phenomenon.
- Contentment: A calm positive emotion experienced after successfully resolving a knowledge gap or reaching a satisfactory explanation.
- Frustration: A negative emotion experienced when repeated attempts to explain a phenomenon or resolve a knowledge gap remain unsuccessful.
- Boredom: A deactivating negative emotion experienced when students lose interest in the task or no longer see value in continuing to resolve the knowledge gap.

\end{tcolorbox}
\caption{Prompt Template: Classification with LLM from raw text)}
\label{fig:prompt_llm_classification}
\end{figure*}

\begin{figure*}[t]
\centering
\begin{tcolorbox}[
    title={Student Knowledge State Interpretation Guide},
    fonttitle=\bfseries,
    coltitle=white,
    width=\textwidth,
    breakable,
    colback=white,
    colframe=black
]
\small\ttfamily\obeylines

\vspace{2ex}

The following quantitative knowledge graph metrics were computed for this episode across four groups: raw counts, comparison against the Domain KG, comparison against the Episode Research Question KG, and understanding quality of entity mentions.
\vspace{2ex}
Use ALL of these as grounded evidence when assessing whether sensemaking was successful:

\textbf{\{metrics\_block\}}

Interpretation guide:
\textbf{[RAW COUNTS]}
- Student Entities:          total concepts the students mentioned in the dialogue
 - Student Triples:           total relations the students expressed
- Correct Triples:           student relations that are scientifically valid (higher = stronger reasoning)
- Incorrect Triples:         student relations that contradict the domain ontology (higher = more misconceptions)
- Question Entities:         concepts present in the episode research question
- Question Triples:          relations present in the episode research question

\textbf{[vs DOMAIN ONTOLOGY]}
- Domain Concept Recall:     \% of domain concepts mentioned (higher = broader curricular coverage)
- Domain Concept Precision:  \% of student concepts that are in the domain (lower = more off-topic mentions)
- Domain Concept F1:         harmonic mean of concept recall and precision
- Domain Relation Recall:    \% of domain relations expressed (higher = richer relational understanding)
- Domain Relation Precision: \% of student relations that exist in the domain
- Domain Relation F1:        harmonic mean of relation recall and precision
- Domain Node Jaccard:       concept set overlap with domain (0=no overlap, 1=identical)
- Domain Edge Jaccard:       relation set overlap with domain
- Domain GED Ratio:          normalised graph edit distance from domain (0=identical, 1=completely different)
- Domain Cosine Sim:         structural cosine similarity with domain ontology

\textbf{[vs EPISODE RESEARCH QUESTION]}
- Target Question:           the specific research question this episode addresses
- Question Concept Recall:   \% of question-target concepts the students addressed (higher = more on-task)
- Dialogue Question Focus:   \% of student concepts centred on the question targets (lower = more off-task drift)
- Question Relation Recall:  \% of question-target relations the students expressed
- Question Node Jaccard:     concept overlap between student KG and question KG
- Question Edge Jaccard:     relation overlap between student KG and question KG
- Question GED Ratio:        normalised edit distance from question KG
- Question Cosine Sim:       structural cosine similarity with question KG
- Question Concepts Matched: raw count of question concepts the students addressed
- Question Concepts Missing: raw count of question concepts the students never mentioned

\textbf{[UNDERSTANDING QUALITY]}
- Correct Understanding:     entity mentions where the student demonstrably understood the concept
- Partial Understanding:     entity mentions where understanding was incomplete or approximate
- Incorrect Understanding:   entity mentions where the student expressed a clear misconception
- Unknown Understanding:     entity mentions where context was insufficient to judge

\end{tcolorbox}
\caption{Prompt Template: KG metrics interpretation guide)}
\label{prompt:prompt_KGmetrics}
\end{figure*}

\begin{figure*}[t]
\centering
\begin{tcolorbox}[
    title={LLM Classification from raw text (part 2)},
    fonttitle=\bfseries,
    coltitle=white,
    width=\textwidth,
    breakable,
    colback=white,
    colframe=black
]
\small\ttfamily\obeylines

\#\#\# sense\_making\_steps  
STEP 1: This stage is where a knowledge gap or inconsistency is identified. Step 1 is present when at least one student notices that something is unclear, contradictory, unexpected, or inconsistent with what they know, observe, or expect while attempting to articulate a phenomenon using their assembled knowledge framework.

Typical indicators include:
- confusion or uncertainty about the phenomenon,
- surprise at an unexpected result,
- questions such as why, how, or what does this mean,
- statements that the data or observation does not fit their expectation
These indicators are illustrative rather than exhaustive. Consider any other dialogue evidence that is consistent with the definition of this step.

STEP 2: Step 2 is present when students actively try to explain the phenomenon or resolve the identified knowledge gap. They may propose, compare, revise, question, or reject explanations while evaluating them using scientific ideas, observations, or evidence.

Typical indicators include:
- proposing or revising explanations
- comparing alternative explanations
- questioning or critiquing ideas
- returning to the same unresolved question (vexing question)
- using observations, data, or scientific principles to support an explanation
These indicators are illustrative rather than exhaustive. Consider any other dialogue evidence that is consistent with the definition of this step.

The quality of this stage is directly proportional to the presentation of a strong and well-supported claim, evidence, and evidence.
A strong explanation includes:
- a clear claim (a statement or conclusion that directly answers the focal question or describes the phenomenon under investigation)
- relevant evidence (appropriate and sufficient scientific data, observations, or arguments that support the claim, rather than personal beliefs or opinions)
- rationale (logical justification that connects the evidence to the claim, which can include relevant scientific principles, mechanisms, or theories)

STEP 3: Step 3 is present only in successful sensemaking when students reach a shared explanation that resolves the earlier gap or inconsistency, at least from their own perspective.

A sensemaking process is considered \"successful\" if all three steps (Step 1, Step 2, and Step 3) are met.
A sensemaking process is considered \"unsuccessful\" if Step 1 is met, but Step 2 is not met or is ineffective, and Step 3 is not met.

\vspace{2ex}
\#\#\# Instructions

1. Based on the student dialogues and sensemaking definition, indicate whether this group has successfully or unsuccessfully completed sensemaking.
2. Provide the main question of the episode in a single sentence (max 20 words).
3. Using the ``Knowledge gap types'' definition, identify the observed knowledge gap type (``none'', ``conceptual'', ``procedural'', or ``epistemic'').
4. If a knowledge gap exists, provide the claim, evidence, and rationale the students produced - one sentence each, max 25 words per field. If no gap exists, use null for all three fields.
5. Choose the two most prominent emotions from this list in order: {emotion\_lists}.

\vspace{2ex}
\#\#\# Output: 
Output ONLY a single valid JSON object in exactly this format - no markdown fences, no extra text:
\{
  sense\_making: successful or unsuccessful,
  main\_question: max 20 words,
  gap\_type: none, conceptual, procedural or epistemic,
  claim: max 25 words or null,
  evidence: max 25 words or null,
  rationale: max 25 words or null,
  emotions: [PrimaryEmotion, SecondaryEmotion]
\}

\vspace{2ex}
\#\#\# Main Task \\
   main\_task =  Analyze the following episode and output the desired JSON:\\ Episode:\textbackslash n\{input\_text\}

\vspace{2ex}
\#\#\# Prompt Variants \\ 

- Variant 1: system role +  sense making steps  + instructions  + main task 
- Variant 2: system role +  sense making steps  + emotion definitions + instructions  + main task 
- Variant 3: system role +  sense making steps  + KG types + instructions  + main task 
- Variant 4: system role +  sense making steps  + KG types + emotion definitions + instructions  + main task 

\end{tcolorbox}
\caption{Prompt Template: Classification with LLM from raw text (part2))}
\label{fig:prompt_llm_classification_part2}
\end{figure*}

\begin{figure*}[ht!]
\centering

\subfloat[SM5\label{fig:SM5}]
{\includegraphics[width=0.28\textwidth]{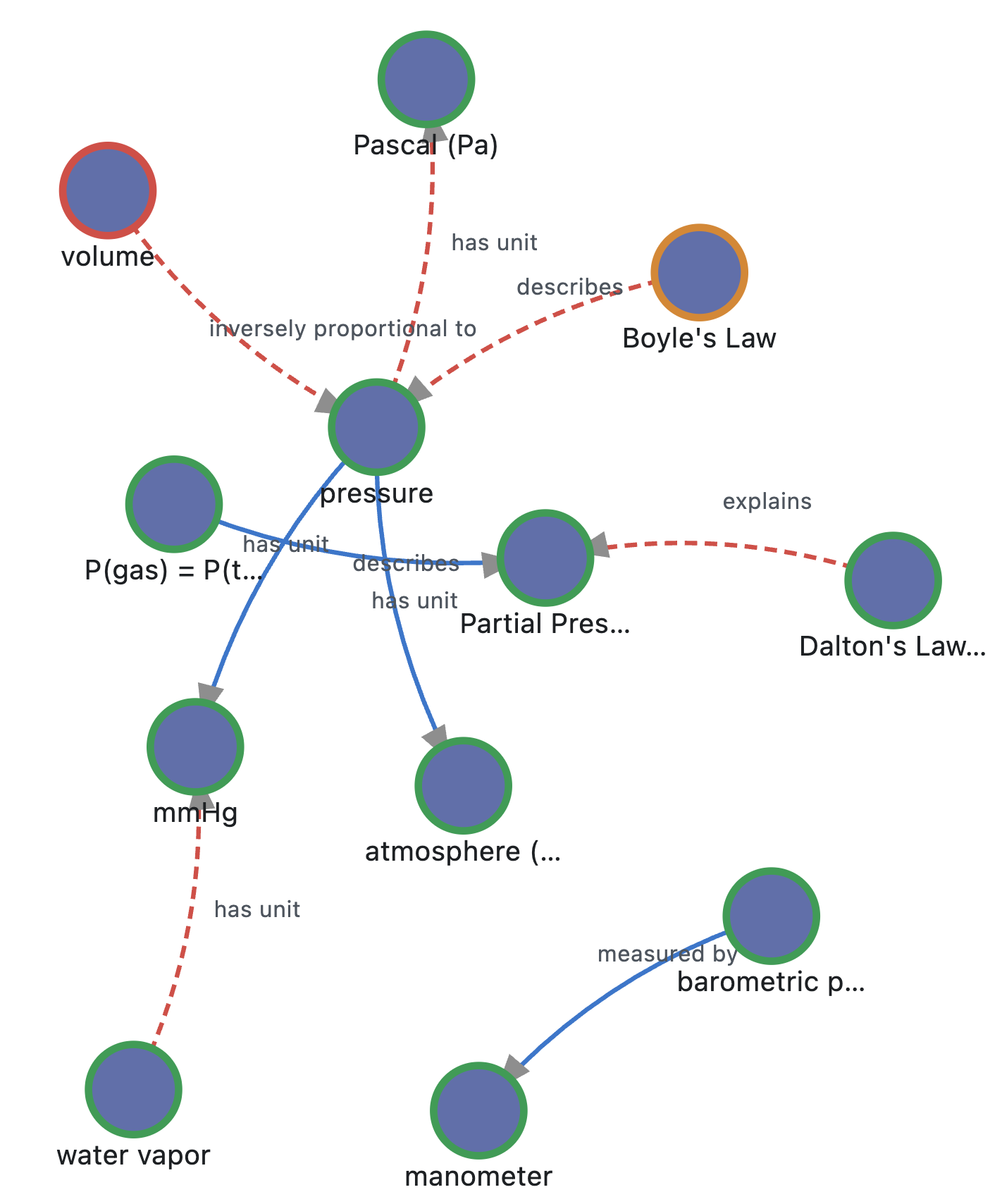}}
~
\subfloat[USM4\label{fig:USM4}]
{\includegraphics[width=0.28\textwidth]{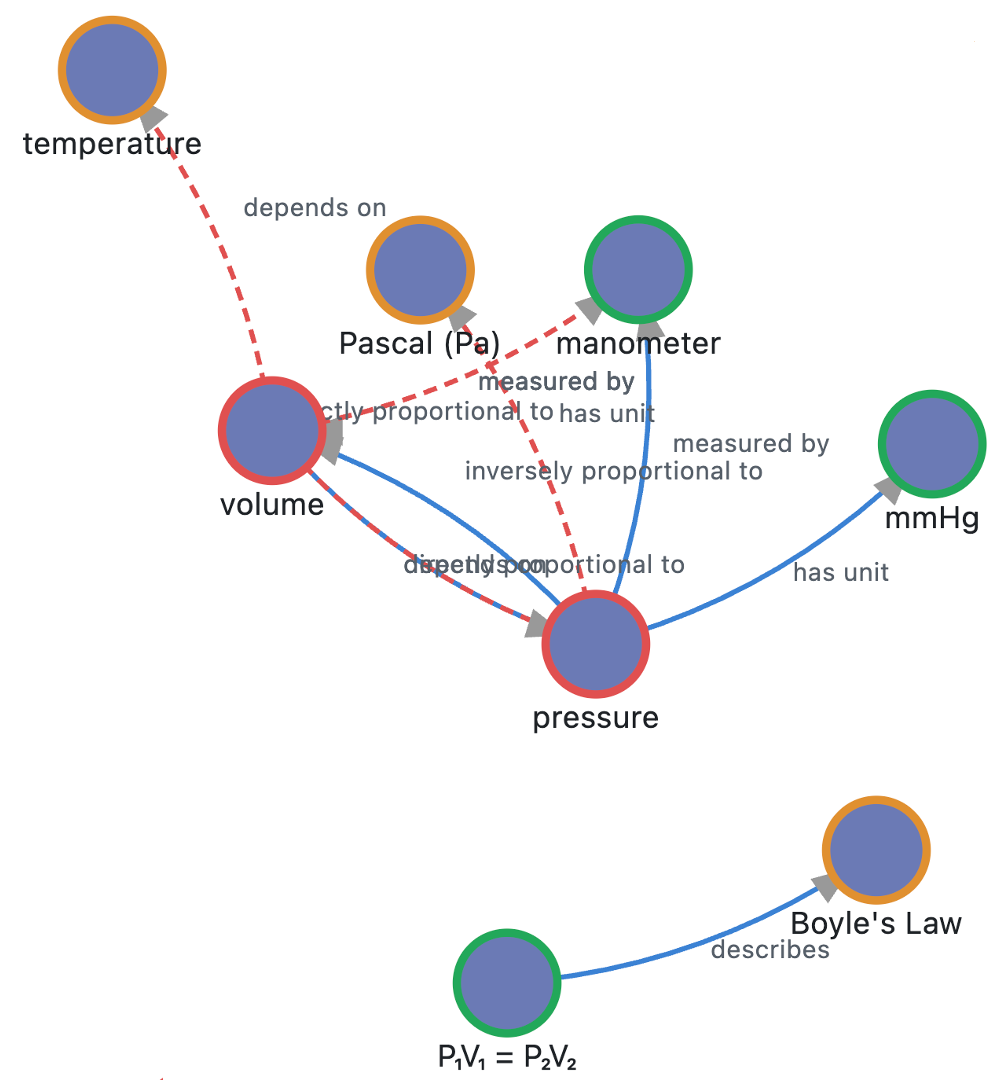}}

\vspace{-1mm}	
\caption{ Student KGs from successful and unsuccessful sense-making episodes.}
\label{fig:studentKGs_samples}
\vspace*{-3mm}
\end{figure*}

\subsection{}{KG Evaluation Metrics}
\label{app:kg_eval_metrics}
\textbf{Raw Counts Metrics}
\begin{itemize}[noitemsep]
    \item Total entities the LLM spotted in the dialogue (nodes in the student KG)
    \item Total relations extracted from the dialogue (edges in the student KG)
    \item Triples with correct flag is true 
    \item Triples with incorrect flag
    \item Entities identified in the episode's research question KG
    \item Relations identified in the episode's research question KG
\end{itemize}

\textbf{Student KG vs. Domain Ontology  Metrics}
\begin{itemize}[noitemsep]
    \item \% of domain concepts the students mentioned; higher = broader curricular coverage
    \item \% of student concepts that are in the domain; lower = more off-topic or hallucinated mentions 
    \item harmonic mean of entity recall and precision
    \item \% of domain relations the students expressed (edge precision)
    \item \% of student relations that exist in the domain (edge recall)
    \item harmonic mean of edge recall and precision
    \item Structural cosine similarity; treats node and edge counts as a vector
\end{itemize}

\textbf{Student KG vs. Episode Research Question metrics}

\begin{itemize}[noitemsep]
    \item \% of question-target concepts that students discussed; 100\% = all question concepts addressed
    \item \% of the student's total concepts that are question-relevant; low = students drifted off-task
    \item Node overlap ratio between student and question KGs
    \item Edge overlap ratio between student and question KGs
    \item Structural similarity between student KG and question KG
    \item How many question-target concepts were actually addressed
\end{itemize}

\textbf{Overall Label Quality}
\begin{itemize}[noitemsep]
    \item correct - Entities where the student demonstrably understood the concept correctly	Higher = stronger conceptual command
    \item partial - Entities where understanding was incomplete or only approximately right	Signals productive struggle or surface engagement
    \item incorrect - Entities where the student expressed a clear misconception	Higher = more conceptual errors present
    \item unknown - Entities where context was insufficient to judge understanding	Usually very brief or implicit mentions
\end{itemize}

\subsection{Fine-grained Breakdown for Prompt variation and knowledge gap type classification}

Table~\ref{tab:basemodels_prompts} reports baseline model performance across prompt variations, while Table~\ref{tab:KGT_results} presents knowledge-gap type prediction results for all Qwen3-32B configurations — the most challenging classification task in our evaluation.

\begin{table}[]
\centering
\caption{Sensemaking variation prediction with prompt variants}
\label{tab:basemodels_prompts}
\resizebox{\columnwidth}{!}{%
\begin{tabular}{lllll}
\toprule
\textbf{Model} & \textbf{V1} & \textbf{V2} & \textbf{V3} & \textbf{V4} \\
Gemma No-Think & 0.172 & \textbf{0.222} & 0.172 & 0.172 \\
Gemma Thinking & \textbf{0.455} & 0.371 & 0.416 & 0.432 \\
Qwen3 No-Think & \textbf{0.358} & 0.188 & 0.358 & 0.358 \\
\textbf{Qwen3 Thinking} & \textbf{0.416} & \textbf{0.021} & \textbf{0.511} & 0.358 \\ \bottomrule
\end{tabular}%
}
\end{table}

\begin{table}[h]
\centering
\resizebox{\columnwidth}{!}{%
\begin{tabular}{rcclll}
\toprule
\textbf{Model Config} & \textbf{Best Prompt} & \textbf{Conceptual} & \textbf{Procedural} \\ \midrule

No-Thinking, single-turn &  0\%  & 100\% (all)\\
No-Thinking, two-turn &   30\% (V1,2,4) & 100\% (all) \\ 
Thinking, single-turn   & 30\% (V3) & 100\% (all) \\ 
Thinking, two-turn  & 50\% (V1) & 100\% (all)\\ \midrule
\textbf{KG-informed Variants } & & &  \\
No-Thinking, single-turn  &  40\% (V3) & 100\% (all) \\
No-Thinking, two-turn  &  40\% (V3)& 100\% (all) \\
Thinking, single-turn  &  40\%  (V3)& 100\% (all)   \\
Thinking, two-turn  &  50 \% (V1, V3) & 100\% (all)\\
\bottomrule
\end{tabular}%
} 
\caption{Qwen3-32B : Sensemaking Variation Prediction}
\label{tab:KGT_results}
\end{table}

\end{document}